\documentclass[letterpaper]{article} % DO NOT CHANGE THIS
\usepackage[preprint]{aaai2027}  % arXiv (de-anonymized): shows authors, drops review-only copyright box. Use [submission] for AAAI review.
\usepackage[hyphens]{url}  % DO NOT CHANGE THIS
\usepackage{graphicx} % DO NOT CHANGE THIS
\usepackage{natbib}  % DO NOT CHANGE THIS AND DO NOT ADD ANY OPTIONS TO IT
\usepackage{caption} % DO NOT CHANGE THIS AND DO NOT ADD ANY OPTIONS TO IT
\usepackage{algorithm}
\usepackage{algorithmic}
\usepackage{newfloat}
\usepackage{listings}
\usepackage{multirow}
\DeclareCaptionStyle{ruled}{labelfont=normalfont,labelsep=colon,strut=off} % DO NOT CHANGE THIS
\floatstyle{ruled}
\newfloat{listing}{tb}{lst}{}
\floatname{listing}{Listing}
\usepackage{booktabs}
\usepackage{amsmath}
\usepackage{amssymb}
\usepackage{subcaption}

\title{REVEAL: A Rubric-Guided Agent for Explicit Evidence Sufficiency Verification in Long-Video Question Answering}
\author{
    Caijun Yan\textsuperscript{\rm 1}\equalcontrib,
    Yang Zhou\textsuperscript{\rm 1}\equalcontrib,
    Meixing Shi\textsuperscript{\rm 1},
    Haoran Sun\textsuperscript{\rm 2},
    Yichen Li\textsuperscript{\rm 2},
    Yuxiang Cai\textsuperscript{\rm 1}\corresponding,
    Yankai Jiang\textsuperscript{\rm 2}\corresponding
}
\affiliations{
    \textsuperscript{\rm 1}Zhejiang University\quad
    \textsuperscript{\rm 2}Shanghai AI Laboratory\\
    yancaijun@zju.edu.cn, caiyuxiang@zju.edu.cn
}

\begin{document}

\maketitle

\begin{abstract}
Recently, retrieval-augmented and memory-augmented methods have emerged as two promising paradigms for long-video question answering. However, existing methods typically rely on rigid, fixed-length temporal chunking (e.g., 10s) and static offline memory banks, which not only fragment coherent continuous events but also fail to adapt during real-time reasoning. Moreover, whether using multi-scale summaries or multimodal knowledge graphs, current approaches prioritize retrieval relevance while overlooking evidence sufficiency, often stopping to answer once only semantically relevant clues are retrieved, even when key temporal, causal, or fine-grained action evidence is still missing. To tackle these challenges, we propose REVEAL, a rubric-guided agent framework. As a foundation, we introduce an adaptive visual-similarity-based preprocessing pipeline that groups visually coherent adjacent frames into natural event units to construct an offline-online video memory---capturing global video context offline while dynamically maintaining question-conditioned memory online. Built upon this structured memory, REVEAL uses an automatically constructed rubric library to explicitly verify whether retrieved evidence satisfies sufficiency criteria, pinpoints missing clues upon verification failure, and directs targeted re-retrieval for complementary information. Without any extra training, REVEAL consistently outperforms both closed-source and open-source state-of-the-art methods across extensive experiments. These results show that explicitly verifying evidence sufficiency, rather than stopping at semantic relevance, retrieves the decisive clues that prior methods miss and yields more reliable long-video reasoning.
\end{abstract}

% Uncomment after de-anonymized camera-ready release.
% \begin{links}
%     \link{Code}{https://anonymous.4open.science/r/rubric-video-repair}
% \end{links}

\section{Introduction}

\begin{figure*}[t]
\centering
\includegraphics[width=\textwidth]{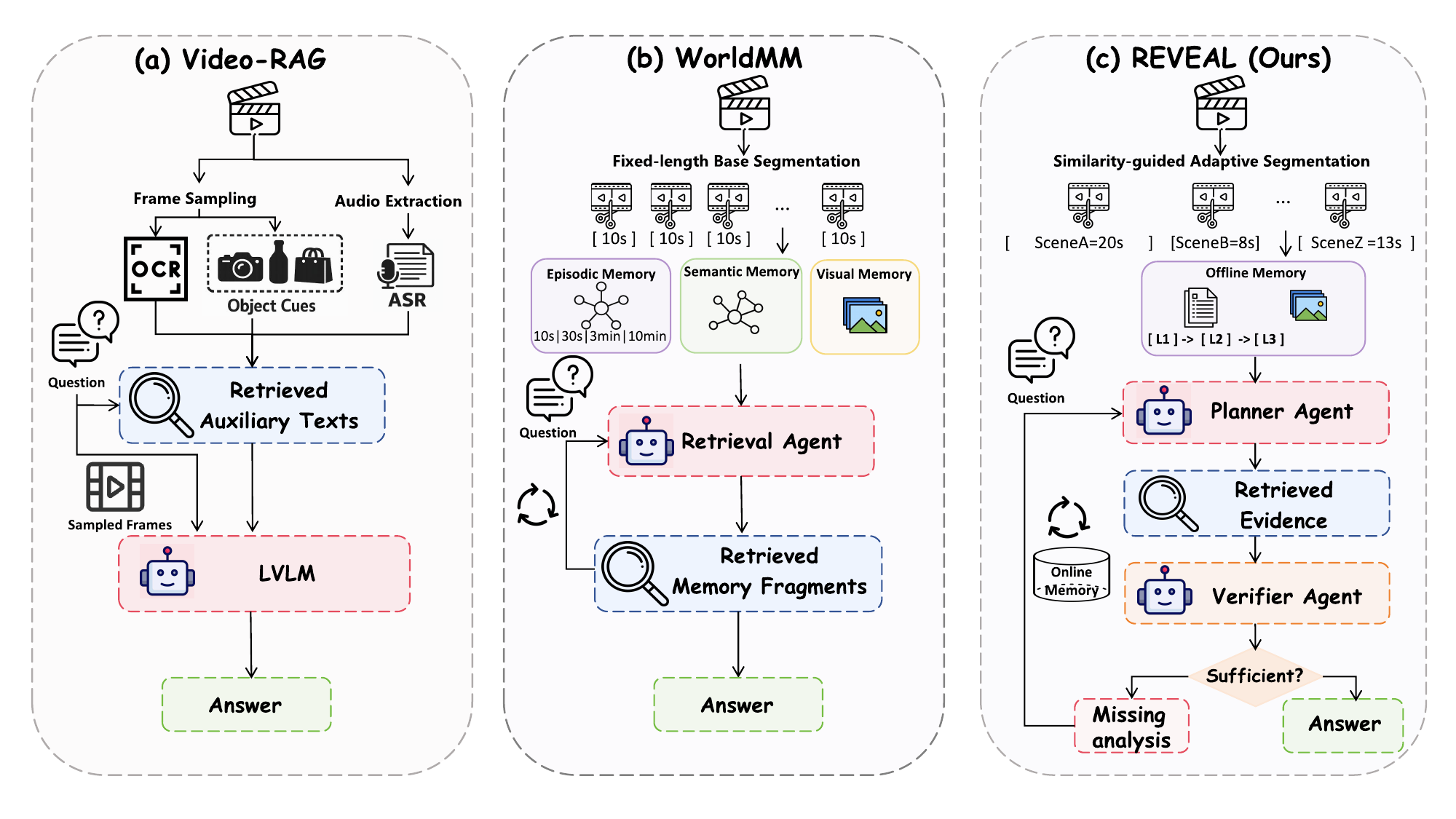}
\caption{Comparison of retrieval-augmented, memory-augmented, and REVEAL paradigms for long-video QA. REVEAL adds adaptive video memory and explicit evidence-sufficiency verification to the inference loop.}
\label{fig:comparison}
\end{figure*}

Unlike short-video question answering, where the relevant context is confined to several minutes, long-video QA must locate and reason over evidence sparsely distributed across hours of video~\citep{fu2024videomme,wang2024lvbench,zhou2024mlvu,wu2024longvideobench,tian2025egor1,chen2024cgbench,kurpath2025longshotbench}. Direct video LLMs expand the temporal context window by employing sparse frame sampling or visual token compression~\citep{video_chatgpt,video_llava,longvu}. While effective at capturing global context, compressing hours of video into a limited token budget discards the fine-grained actions, state transitions, and small textual clues crucial for reasoning.

To circumvent encoding full-length videos, retrieval-augmented methods index sampled clips or text transcripts and retrieve query-relevant subsets during inference~\citep{lewis2020rag,video_rag} (Figure~\ref{fig:comparison}(a)). Memory-augmented approaches go a step further by constructing multi-granularity memory banks to support multi-round retrieval and reasoning~\citep{moviechat,ma_lmm,fan2024videoagentmemory,worldmm,tian2025egor1,chen2025lvagent,wang2025avp} (Figure~\ref{fig:comparison}(b)). Despite their progress, these paradigms exhibit clear structural and operational limitations. Structurally, many existing RAG and memory frameworks rely on rigid, fixed-duration temporal chunking (e.g., 10s, 30s, or 1min) to index videos or build hierarchical abstractions. This arbitrary slicing fragments coherent, variable-length event units across segment boundaries. Operationally, retrieval and stopping are driven by semantic relevance rather than evidence sufficiency, frequently leading agents to answer based on segments that are topically related to the question but lack decisive causal, temporal, or fine-grained action details. Furthermore, during multi-turn retrieval, existing frameworks rarely track past retrieval quality or explicit missing clues, often trapping the agent in redundant, unguided searches.

Driven by these limitations, we propose REVEAL, a novel agentic MLLM framework. REVEAL replaces fixed temporal units with visually coherent, variable-length L1 segments and summarizes them into L2 event timelines and L3 global narratives (Figure~\ref{fig:comparison}(c)). This reusable offline memory preserves local evidence and long-range context, while a question-conditioned online memory tracks admitted evidence, attempted queries, retrieval quality, and the verifier's latest evidence gap. Compared with a static memory bank, this separation lets inference retain and update the state needed for targeted repair retrieval.

Additionally, REVEAL replaces implicit stopping with rubric-guided evidence-sufficiency verification. An automatically constructed rubric library is induced from contrastive pairs of decisive and misleading evidence. At inference time, the verifier selects criteria relevant to the question and scores whether the accumulated evidence is sufficient to judge the answer options. If the threshold is met, the verifier triggers the answerer to generate the final answer; otherwise, the verifier produces a concrete missing-evidence analysis. The planner uses this diagnosis, search history, and retrieval-quality feedback to issue focused queries or switch to uniform timeline sampling when targeted directions are exhausted. Thus, retrieval is controlled by what is still needed, rather than by semantic similarity alone.

Our contributions are summarized as follows:
\begin{itemize}
    \item We are the first to explore rubric-based criteria as an inference-time mechanism for long-video QA agents. Rather than using rubrics only to assess model outputs or optimize training objectives, we show that they can serve as interpretable criteria that guide agent decision-making during inference.

    \item We design REVEAL, a rubric-guided long-video QA agent that realizes this paradigm by integrating an offline-online video memory built from similarity-grouped scene segments with a verifier that evaluates evidence sufficiency, diagnoses missing evidence, and guides targeted retrieval. We further introduce an automatic rubric construction pipeline that derives rubrics from contrastive evidence pairs rather than manual authoring.

    \item REVEAL is training-free and, with an open backbone, achieves state-of-the-art performance across multiple long-video QA benchmarks, surpassing strong open-source and commercial baselines.
\end{itemize}
\section{Related Work}

\subsection{Long Video Understanding}

Long-video understanding requires reasoning over sparse evidence distributed across extended temporal contexts, which is difficult to feed directly into MLLMs under visual-token and context-length constraints. Existing approaches therefore reduce video input through sparse frame sampling, keyframe selection, or visual-token compression~\citep{video_chatgpt,video_llava,longvu,qwen2_vl}, or avoid full-video encoding by indexing videos into external memories and retrieving a small set of relevant clips or memory items for each query~\citep{lewis2020rag,video_rag,fan2024videoagentmemory,worldmm,tian2025egor1}. Recent memory-based systems further organize videos into structured representations, such as captions, hierarchical multi-scale summaries, and entity knowledge graphs, while agentic approaches incorporate planning, tool use, and multi-round retrieval~\citep{moviechat,ma_lmm,wang2024videoagentlongform,yang2024vca,chen2025lvagent,wang2025avp,yao2023react,liu2025videomind,yeo2025gcagent}. These methods substantially improve access to relevant video information, but relevance alone does not guarantee sufficient evidence for answering a question. Moreover, decisions about whether enough evidence has been collected are typically based on implicit model judgments~\citep{asai2023selfrag,jiang2023flare,yan2024crag}.

\subsection{Rubric-Based Methods}

Rubrics decompose subjective assessments into explicit, verifiable criteria, separating evaluation standards from the judging model. This makes LLM-as-a-judge evaluations more interpretable and has been widely applied to open-ended tasks~\citep{liu2023geval,zheng2023judging,kim2023prometheus,arora2025healthbench,waheed2025videojudge,yang2026healthscore}, with recent studies exploring automatic rubric construction at scale~\citep{liu2026rubricssurvey,li2026rubrichub,fan2025sedareval,gao2026qworld,qi2026rift}. Beyond evaluation, rubrics have also been adopted as reward signals in reinforcement learning, where LLM-based graders provide fine-grained feedback for tasks without explicit ground-truth answers~\citep{schulman2017ppo,liu2025openrubrics,zhou2025ruscarl,zhang2026chasingtail}. However, existing applications primarily use rubrics after an answer is produced, either to evaluate outputs or optimize training objectives. We instead move rubrics into the inference loop of a long-video agent, where they serve as interpretable criteria for assessing whether retrieved evidence is sufficient and identifying unmet criteria that guide subsequent retrieval.
\section{Methodology}

\begin{figure*}[t]
\centering
\includegraphics[width=\textwidth]{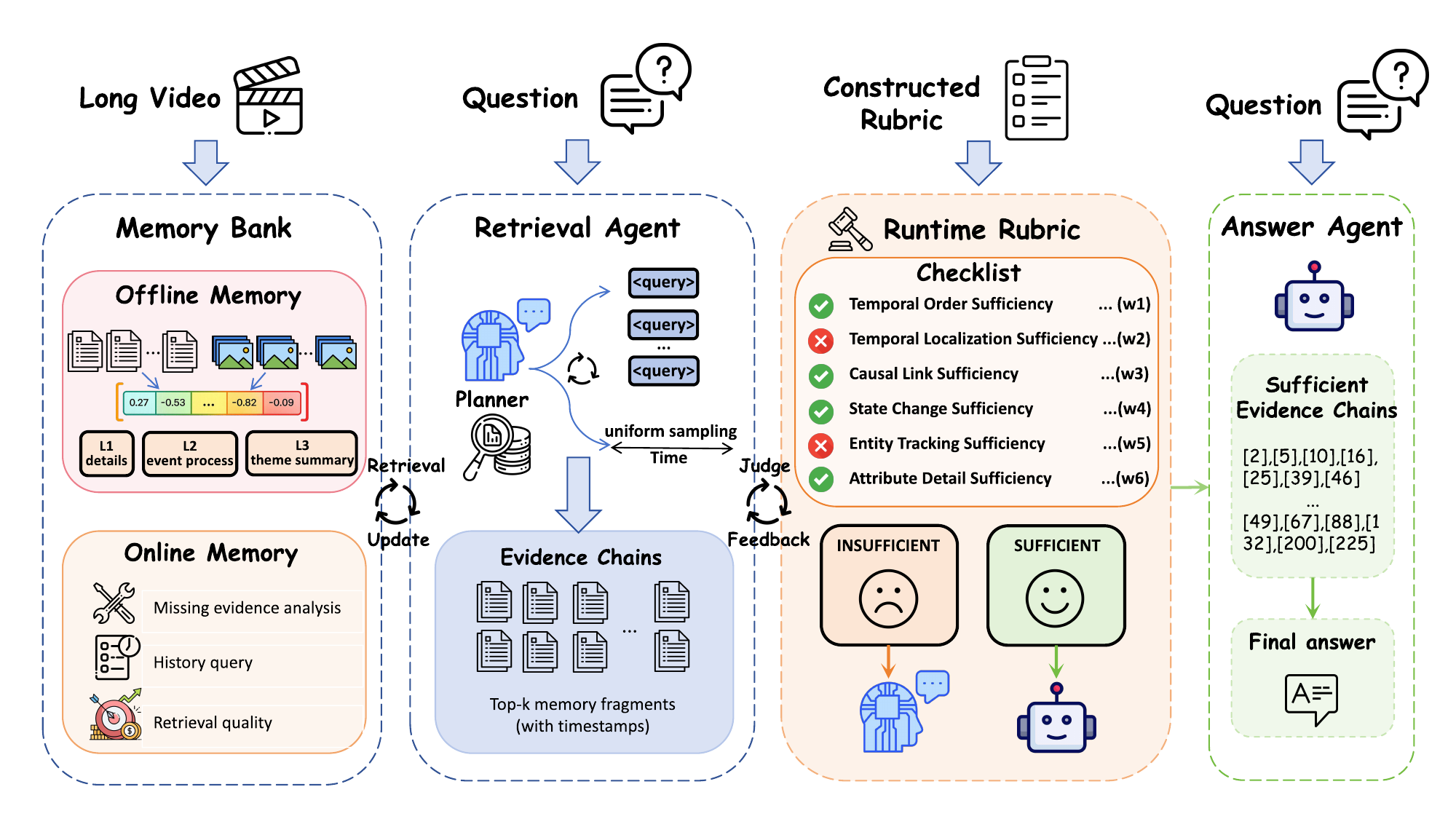}
\caption{Overview of REVEAL. A hierarchical offline memory supports multi-scale retrieval; the online memory tracks evidence gaps and search history; and the planner and rubric-guided verifier iteratively retrieve and validate evidence before answering.}
\label{fig:overview}
\end{figure*}

As illustrated in Figure~\ref{fig:overview}, REVEAL introduces rubric-guided evidence-sufficiency verification into long-video QA, retrieving additional information whenever the current evidence does not meet the question's requirements. It comprises three components: (1) an \emph{offline-online memory framework} separating reusable video representations from question-dependent evidence accumulation (Section~\ref{sec:memory}); (2) a \emph{rubric construction pipeline} that derives sufficiency criteria from decisive and misleading evidence (Section~\ref{sec:rubric}); and (3) a \emph{rubric-guided retrieve--verify--answer loop} that iteratively retrieves and verifies evidence before answering (Section~\ref{sec:loop}). A shared VLM serves as planner, verifier, and answerer through role-specific prompts, without supervised fine-tuning or reinforcement learning.

\subsection{Offline-Online Memory Framework}
\label{sec:memory}

Long-video QA involves two distinct forms of memory: a stable representation of what happened in the video and a transient reasoning state that records what has been established for the current question. Conflating them either repeats expensive video processing for every query or leaves the retrieval process unaware of its previous attempts. REVEAL therefore factorizes memory into a persistent offline memory $\mathcal{M}$ and a question-conditioned online memory $\mathcal{O}_t$. The former is constructed once and shared by all questions about the same video; the latter is instantiated per question and evolves with the evidence-acquisition process.

\noindent\textbf{Offline memory.}
To preserve fine visual detail without forcing every query to search an hour-scale sequence at frame level, we construct a three-level temporal hierarchy that balances local fidelity and long-range abstraction. At its finest level, adjacent 1-fps frames are grouped by visual similarity into coherent, variable-length L1 segments, avoiding the splitting of a continuous event across the arbitrary fixed windows used by prior work. An L1 item is represented as $m_i^{(1)}=(t_i^s,t_i^e,x_i^{(1)},\mathbf{h}_i^{(1)},\mathbf{v}_i)$, where $(t_i^s,t_i^e)$ anchors the event in time, $x_i^{(1)}$ describes its multimodal content, $\mathbf{h}_i^{(1)}$ is its textual embedding, and $\mathbf{v}_i$ preserves a representative visual embedding. Consecutive L1 narratives are then summarized bottom-up into L2 event timelines and an L3 global narrative, yielding
\begin{equation}
\mathcal{M}=\mathcal{M}_{\mathrm{L1}}\cup\mathcal{M}_{\mathrm{L2}}\cup\mathcal{M}_{\mathrm{L3}}.
\end{equation}
The three levels expose complementary retrieval units: L1 preserves local actions and appearances, L2 captures event progression, and L3 supplies video-level context. BGE-M3~\citep{chen2024bgem3} encodes the narrative at every level into $\mathbf{h}_i^{(\ell)}$ for text retrieval, while L1 additionally retains $\mathbf{v}_i$ for access to the underlying perceptual evidence. A query can thus retrieve a precise moment, a temporally extended event, or a global narrative through the same memory interface, which prevents global questions from being answered from isolated local matches. Construction and indexing details are provided in the supplementary material.

\noindent\textbf{Online memory.}
Offline retrieval alone has no notion of whether a returned item is new, redundant, or useful for resolving the remaining uncertainty. The online memory supplies this missing query-level state. At round $t$, it is defined as
\begin{equation}
\mathcal{O}_t=(\mathcal{E}_t,\mathcal{V}_t,\mathcal{H}_t,\mathcal{N}_t,\mathcal{A}_t),
\end{equation}
where $\mathcal{E}_t$ is the deduplicated evidence cache, $\mathcal{V}_t$ contains visited memory IDs, $\mathcal{H}_t$ records attempted history queries, $\mathcal{N}_t$ indicates whether each attempt contributed new evidence, and $\mathcal{A}_t$ is the verifier's latest analysis of the missing evidence. These fields jointly encode three complementary signals: the current knowledge state, the explored search space, and the unresolved information need. The planner observes the compact feedback tuple $(\mathcal{H}_t,\mathcal{N}_t,\mathcal{A}_t)$ instead of verbose criterion-level scores, so verification results guide retrieval and unproductive searches are recognized explicitly. The update contract is detailed in the supplementary material.

\subsection{Rubric Construction Pipeline}
\label{sec:rubric}

\begin{figure}[t]
\centering
\includegraphics[width=\columnwidth]{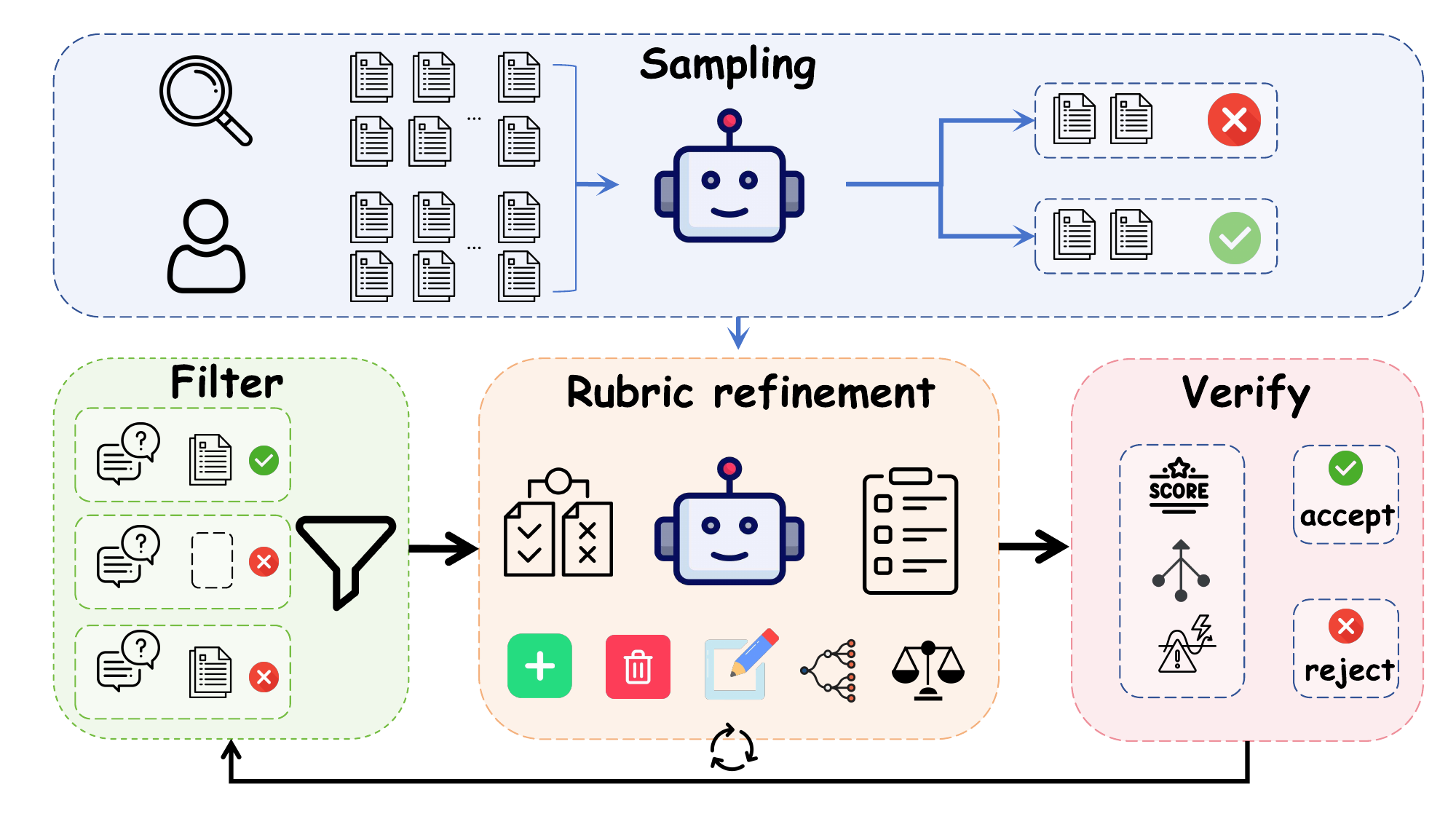}
\caption{Contrastive rubric construction pipeline. A teacher labels sufficient and misleading evidence; answer-dependent filtering yields contrastive pairs; a proposer refines criteria; and held-out errors form the next working set.}
\label{fig:Rubric_construction}
\end{figure}

Semantic relevance is an inadequate stopping criterion: a retrieved clip may mention the correct entities yet omit the temporal order, state change, or discriminative detail required by the question. We instead formulate stopping as an evidence-sufficiency judgment governed by an explicit rubric. Manually authoring effective rubrics presents a fundamental trade-off: high-level, generic criteria fail to capture the subtle failure modes that alter an answer, whereas exhaustively crafting fine-grained rules for specific error cases imposes prohibitive expert annotation costs. Our key observation is that sufficiency can be learned from the boundary between decisive and deceptively relevant evidence. REVEAL therefore constructs a reusable rubric library offline from contrastive evidence pairs, following the pipeline in Figure~\ref{fig:Rubric_construction}. This grounds each criterion in the decision boundary between sufficient evidence that leads to correct predictions and deceptive, insufficient evidence that misleads the model into incorrect answers. The rubric library is constructed exclusively from the training set, with no exposure to the test set.

Given a question $q$ with ground-truth answer $y$, a teacher identifies a decisive evidence set $E^+$ and a topically relevant yet non-decisive set $E^-$. To ensure each pair exposes a genuine sufficiency boundary, we keep a candidate only when the answer model $A$ satisfies:
\begin{equation}
A(q,E^-) \neq y, \quad A(q,\varnothing) \neq y, \quad A(q,E^+) = y.
\end{equation}
These conditions guarantee the validity of each evidence pair: $A(q,\varnothing)\neq y$ prevents the model from answering via blind guessing or background memory without visual context; $A(q,E^+)=y$ confirms that $E^+$ is genuinely sufficient; and $A(q,E^-)\neq y$ ensures that $E^-$ lacks the key detail rather than accidentally leaking the answer. The retained pairs isolate the fine-grained discriminative properties that transition an answer from unsupported to correct. Next, an LLM-based proposer abstracts these properties into reusable, weighted criteria with operational score boundaries ($0$ for missing evidence, $0.5$ for partially sufficient, and $1$ for sufficient). Rather than relying on single-pass generation, we iteratively refine these criteria against held-out contrastive cases using validation error feedback. The refined criteria are consolidated offline into a single unified rubric pool; at inference, the verifier selects a question-specific subset from this fixed pool rather than authoring new rules per question, converting recurring retrieval failure modes into an explicit and inspectable sufficiency test. Complete algorithmic details and example rubrics are provided in the supplementary material.

\subsection{Rubric-Guided Retrieve--Verify--Answer Loop}
\label{sec:loop}

Given the offline memory $\mathcal{M}$ and the unified rubric pool, REVEAL alternates between evidence acquisition and sufficiency verification. The planner translates the current information gap into a retrieval action, while the verifier determines whether the accumulated evidence can discriminate among the answer options.

\paragraph{Planner.}
The planner serves two different roles over the course of inference. In the initial round, it decomposes the question and options into 2--4 short, self-contained queries, improving recall when the answer depends on multiple entities or events. In each subsequent repair round, it conditions on the accumulated evidence and online state to choose a targeted retrieval action that addresses the diagnosed evidence gap.

At round $t$, the planner policy generates a retrieval action $a_t$ based on the current online memory state:

\begin{equation}
a_t = P(q, \mathcal{O}_{t-1}).
\end{equation}

Specifically, $a_t$ specifies the retrieval strategy over the offline memory hierarchy. On one hand, targeted retrieval leverages the verifier's latest analysis $\mathcal{A}_{t-1}$ to generate focused queries across all three levels ($\mathcal{M}_{\mathrm{L1}}, \mathcal{M}_{\mathrm{L2}}, \mathcal{M}_{\mathrm{L3}}$). On the other hand, exploratory retrieval directly samples unvisited segments ($\mathcal{M}_{\mathrm{L1}} \setminus \mathcal{V}_{t-1}$) when a question requires broader coverage or targeted searches prove insufficient. By conditioning on the feedback tuple $(\mathcal{H}_{t-1}, \mathcal{N}_{t-1}, \mathcal{A}_{t-1})$, the planner explicitly learns what is currently missing ($\mathcal{A}_{t-1}$) and what has previously failed ($\mathcal{H}_{t-1}, \mathcal{N}_{t-1}$), turning each reasoning step into a directed repair of the evidence cache $\mathcal{E}_{t-1}$. Implementation details and prompt templates are provided in the supplementary material.

\paragraph{Rubric-Guided Verification.}
Crucially, REVEAL does \emph{not} generate a new rubric for each question. It maintains a single unified rubric pool built once offline, and at inference the verifier dynamically selects the subset of criteria relevant to the question $q$. Judging against this shared, pre-validated pool, rather than authoring ad hoc per-question rules, keeps criteria reusable and prevents irrelevant rules from affecting the verdict. For each selected criterion $k$, the verifier evaluates the accumulated evidence and assigns a satisfaction score $z_{t,k} \in \{0, 0.5, 1\}$. The overall verification confidence for the question is computed as the weighted aggregation:
\begin{equation}
\rho_t = \frac{\sum_k w_k z_{t,k}}{\sum_k w_k}.
\end{equation}
The loop terminates when $\rho_t \ge \tau$, indicating that the accumulated evidence is sufficient to decide among the options. We emphasize that this certifies evidence \emph{sufficiency}, not answer \emph{correctness}: it asserts that the decisive information has been gathered, not that the resulting prediction is necessarily right. Otherwise, the verifier produces $\mathcal{A}_t$, a concrete description of the missing facts still needed to satisfy the remaining criteria. This design separates two decisions often entangled in agentic retrieval: the scalar score $\rho_t$ controls \emph{whether} to continue, whereas $\mathcal{A}_t$ specifies \emph{what} to retrieve next. In contrast to unconstrained self-reflection~\citep{shinn2023reflexion}, both are grounded in reusable, inspectable criteria. Prompt schemas and deterministic aggregation details are provided in the supplementary material.

\paragraph{Answer Generation.}
Once the verifier accepts the evidence or the retrieval budget is reached, the deduplicated evidence accumulated across rounds is passed to the answerer together with the question and options. The answerer is deliberately isolated from retrieval control: it predicts from the final evidence set, while the planner and verifier determine how that set is acquired. All three roles use the same VLM. If the budget is exhausted below the sufficiency threshold, the answerer returns the best-supported prediction for benchmark compatibility, and the final missing-evidence analysis is retained for diagnosis.

\FloatBarrier
\section{Experiments}

\begin{table*}[t]
\centering
\fontsize{10}{12}\selectfont
\setlength{\tabcolsep}{3.4pt}
\renewcommand{\arraystretch}{1.06}
\begin{tabular}{@{}lccccccc@{}}
\toprule
Method & Size & EgoLifeQA & Ego-R1 & LVBench & Video-MME & LVB & Avg. \\
\midrule
\multicolumn{8}{c}{\textbf{Direct Video MLLMs}} \\
\midrule

% & LongVU
% & 7B
% & 32.6 & 37.3 & 38.3 & 48.2 & 53.5 & 40.0 \\

% & Video-XL
% & 7B
% & 35.2 & 38.7 & 38.5 & 46.3 & 50.4 & 39.8 \\

Video-RTS~\citep{wang2025videorts}
& 7B
& 48.2 & 47.4 & 39.8 & 47.9 & 56.6 & 48.0 \\

Qwen3-VL-32B-Instruct~\citep{bai2025qwen3vl}
& 32B
& 32.2 & 32.3 & 39.4 & 53.1 & 53.5 & 42.1 \\

Qwen3.5-27B~\citep{qwen2026qwen35}
& 27B
& 37.0 & 41.3 & 64.0 & 70.0 & 73.5 & 57.2 \\

Gemini 2.5 Pro~\citep{google2025gemini25}
& $-$
& 46.4 & 46.7 & 57.0 & 55.7 & $-$ & $-$ \\

GPT-5~\citep{openai2025gpt5}
& $-$
& 48.6 & 46.3 & 60.4 & 74.3 & 64.5 & 58.8 \\

\midrule

\multicolumn{8}{c}{\textbf{Retrieval-augmented Methods}} \\
\midrule
Video-RAG~\citep{video_rag}
& 27B
& 46.6 & 47.3 & 47.5 & 66.4 & 52.0 & 52.0 \\

HippoRAG 2~\citep{gutierrez2025hipporag2}
& 27B
& 43.3 & 42.4 & 51.2 & 68.3 & 57.9 & 52.6 \\

Vgent~\citep{shen2025vgent}
& 27B
& 48.7 & 45.7 & 51.9 & 77.0 & 59.6 & 56.6 \\

\midrule

\multicolumn{8}{c}{\textbf{Memory-augmented Methods}} \\
\midrule
M3-Agent~\citep{long2025m3agent}
& 32B
& 53.5 & 52.0 & 49.3 & 55.3 & 47.6 & 51.5 \\

VideoLucy~\citep{zuo2025videolucy}
& 27B
& 42.5 & 51.2 & 50.4 & 68.7 & 52.7 & 53.1 \\

WorldMM~\citep{worldmm}
& 27B
& 49.2 & 50.7 & 48.7 & 58.7 & 57.7 & 53.0 \\

\midrule
\multicolumn{8}{c}{\textbf{Ours}} \\
\midrule
\textbf{REVEAL-27B}
& 27B
& \textbf{64.4} & \textbf{66.7} & \textbf{65.9} & \textbf{79.1} & \textbf{75.7} & \textbf{70.4} \\

\bottomrule
\end{tabular}
\caption{Main performance comparison on five public long-video reasoning benchmarks.
The Video-MME column reports the Long split, and LVB denotes the LongVideoBench validation split; averages are reported only when all five results are available.}
\label{tab:main_results}
\end{table*}

\subsection{Experimental Setup}

\paragraph{Datasets.}
We evaluate REVEAL on five public long-video QA benchmarks using multiple-choice accuracy. Video-MME-L and LVBench test long-form third-person video reasoning, while LongVideoBench evaluates referring and temporal reasoning on its public validation split. EgoLifeQA and Ego-R1 Bench focus on egocentric videos, where answer evidence is sparse and often tied to fine-grained daily activities. Together, these datasets cover both viewpoint regimes and a broad range of video durations and reasoning demands. Per-benchmark statistics, splits, and evaluation protocols are provided in the supplementary material.

\paragraph{Baselines.}
Following the grouping in Table~\ref{tab:main_results}, we compare against three families of systems. \emph{Direct video MLLMs} answer from frames sampled over the video without an explicit retrieval or memory module, spanning both open-source models (Video-RTS and Qwen3-VL-32B-Instruct) and commercial models (Gemini~2.5~Pro and GPT-5). \emph{Retrieval-augmented methods}, which fetch question-specific evidence at inference time, include Video-RAG, HippoRAG~2, and Vgent. \emph{Memory-augmented methods}, which build a persistent video memory before answering, include M3-Agent, VideoLucy, and WorldMM. Together these families span direct perception, external retrieval, and long-term memory, covering the dominant design choices for long-video QA.

\paragraph{Implementation Details.}
We instantiate REVEAL with a single Qwen3.5-27B model, denoted REVEAL-27B, which builds the offline memory and also acts as planner, verifier, and answerer, at temperature zero and without any fine-tuning; its standalone accuracy appears among the direct video MLLMs in Table~\ref{tab:main_results} as a backbone reference. Frames are encoded with SigLIP-Large~\citep{zhai2023siglip}, a CLIP-style~\citep{radford2021clip} contrastive vision-language encoder, and text with BGE-M3~\citep{chen2024bgem3}. We use a single configuration for all five benchmarks, and each reported result is computed from a single evaluation run. For fair comparison, each modular baseline is run with the same Qwen3.5-27B backbone, except M3-Agent, which keeps its released task-trained checkpoint. Further implementation details, together with an efficiency analysis of runtime, token consumption, average number of retrieval rounds, and LLM call cost, are provided in the supplementary material.

\subsection{Main Results}

Table~\ref{tab:main_results} compares REVEAL with recent long-video systems. REVEAL-27B ranks first on all five benchmarks, reaching $64.4$ on EgoLifeQA, $66.7$ on Ego-R1 Bench, $65.9$ on LVBench, $79.1$ on Video-MME-L, and $75.7$ on LongVideoBench. On the two challenging egocentric benchmarks, it exceeds the strongest competitor by $10.9$ and $14.7$ points, respectively.

With the same Qwen3.5-27B backbone, REVEAL raises the direct MLLM average from $57.2$ to $70.4$ and consistently outperforms the reimplemented modular frameworks. This controlled comparison attributes the gain to the framework rather than model scale. With an open backbone, REVEAL is also competitive with, and on the egocentric benchmarks ahead of, strong commercial systems such as GPT-5; since these systems differ substantially in scale, training, and access, we treat such comparisons as indicative rather than strictly controlled. Overall, long-video reasoning depends not only on model capacity, but on distinguishing decisive evidence from merely relevant content.

\begin{table}[t]
\centering
\fontsize{10}{12}\selectfont
\setlength{\tabcolsep}{0.4pt}
\renewcommand{\arraystretch}{1.05}
\begin{tabular}{lcccccc}
\toprule
\multirow{2}{*}{\textbf{Method}}
& \multicolumn{4}{c}{Components}
& \multicolumn{2}{c}{Datasets} \\
\cmidrule(lr){2-5}
\cmidrule(lr){6-7}
& \shortstack{Iter.}
& \shortstack{Offline\\Mem.}
& \shortstack{Online\\Mem.}
& \shortstack{Rubric\\Suff.}
& \shortstack{Video-\\MME(long)}
& LVBench \\
\midrule

$\mathcal{M}_{\mathrm{base}}$
& $-$ & $-$ & $-$ & $-$ & 66.5 & 51.6 \\

$\mathcal{M}_{\mathrm{+iter}}$
& \checkmark & $-$ & $-$ & $-$ & 68.2 & 52.2 \\

$\mathcal{M}_{\mathrm{+OffMem}}$
& \checkmark & \checkmark & $-$ & $-$ & 69.9 & 54.4 \\

$\mathcal{M}_{\mathrm{+OnMem}}$
& \checkmark & \checkmark & \checkmark & $-$ & 71.3 & 55.0 \\

\textbf{Ours}
& \checkmark & \checkmark & \checkmark & \checkmark & \textbf{79.1} & \textbf{65.9} \\
\bottomrule
\end{tabular}
\caption{Ablation on Video-MME(long) and LVBench.
}
\label{tab:ablation_design}
\end{table}

\subsection{Ablation Study}

To isolate each component's contribution, we start from a single-round retrieval baseline ($66.5/51.6$ on Video-MME-L$/$LVBench) and add one component at a time (Table~\ref{tab:ablation_design}). \emph{Iterative retrieval} ($+1.7/+0.6$) lets the agent issue follow-up queries instead of committing to a single retrieval, recovering evidence that an imperfect first query misses; the gain is small because, without an explicit stopping signal, the loop still cannot tell which evidence is missing and often re-searches the same region. \emph{Hierarchical offline memory} ($+1.7/+2.2$) replaces fixed-length chunking with visually coherent variable-length segments and multi-scale summaries, so each retrieved unit spans a complete event rather than a fragment, while the L2/L3 summaries expose the global context that isolated local segments lack; the larger LVBench gain reflects its stronger reliance on long-range temporal structure. \emph{Online memory} ($+1.4/+0.6$) tracks admitted evidence, prior queries, and per-round retrieval quality, which suppresses redundant searches and preserves the state needed to steer the next query toward genuinely new evidence. These three components broaden and organize the evidence pool, but each still stops on relevance, which caps their combined effect. \emph{Rubric-guided sufficiency verification} adds by far the largest jump ($+7.8/+10.9$): by scoring whether the accumulated evidence actually meets the question's criteria, it stops only once the decisive evidence is present and, on failure, names the specific missing clue to drive a targeted repair query. This confirms that the decisive factor is not broader memory coverage but explicitly identifying and repairing unmet evidence requirements, which is exactly where relevance-based stopping fails.

\subsection{Analysis}

\begin{figure*}[t]
\centering
\begin{subfigure}{0.32\textwidth}
\centering
\includegraphics[width=\textwidth]{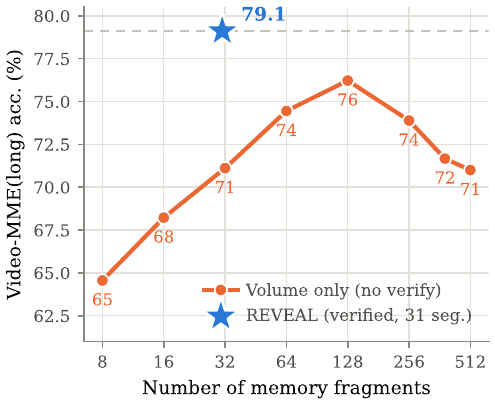}
\caption{}
\label{fig:volume}
\end{subfigure}
\hfill
\begin{subfigure}{0.32\textwidth}
\centering
\includegraphics[width=\textwidth]{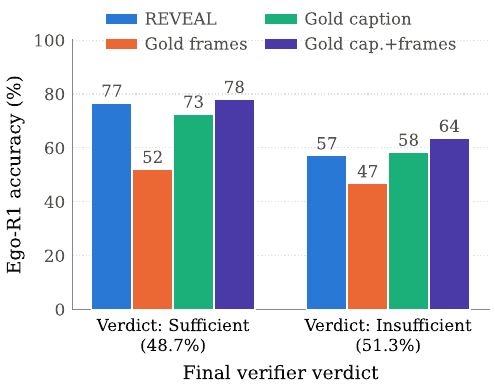}
\caption{}
\label{fig:verifier_oracle}
\end{subfigure}
\hfill
\begin{subfigure}{0.32\textwidth}
\centering
\includegraphics[width=\textwidth]{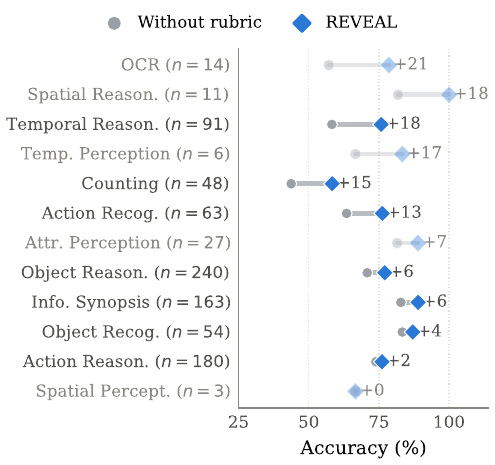}
\caption{}
\label{fig:analysis}
\end{subfigure}
\caption{Analysis of the sufficiency verifier: (a) accuracy vs.\ number of retrieved segments; (b) verifier calibration on Ego-R1; (c) per-type accuracy before/after verification.}
\label{fig:analysis_col}
\end{figure*}

\paragraph{One Rubric Pool Across All Benchmarks.}
Throughout every experiment above, REVEAL relies on a single rubric pool that is constructed once on the training split and then applied unchanged to all five benchmarks, spanning both third-person and egocentric footage. We neither tune nor regenerate criteria per dataset: the same fixed pool drives verification everywhere, and at inference the verifier only selects a question-specific subset from it. The consistent improvements therefore reflect the transferability of one shared set of sufficiency criteria rather than any benchmark-specific engineering.

\paragraph{Effect of Evidence Volume.}
To test whether REVEAL's gains come simply from retrieving more evidence, we disable verification and vary the number of retrieved segments. As shown in Figure~\ref{fig:volume}, accuracy without verification peaks at $76.2\%$ around 128 segments and then declines as extra segments add distraction, whereas REVEAL with sufficiency verification reaches $79.1\%$ using about 31 segments on average. The gap is twofold: verification not only lifts the ceiling by $2.9$ points but reaches it with roughly a quarter of the evidence, so the answerer reasons over a compact, decisive context rather than a relevance-ranked list that dilutes the critical segment. The downward slope beyond 128 segments is itself informative: once the decisive evidence is already present, further relevant-but-redundant segments actively hurt by crowding the context with plausible distractors---precisely the failure mode that an explicit sufficiency test is designed to avoid. This confirms that targeted acquisition, not sheer volume, drives the gain.

\paragraph{Calibration of Sufficiency Verdicts.}
To assess whether the verifier's verdict reflects real evidence sufficiency, we use Ego-R1's \emph{target-time} annotations---the interval containing each answer's supporting evidence---as retrieval-free oracle evidence, and group questions by the final verdict. As shown in Figure~\ref{fig:verifier_oracle}, REVEAL scores $76.7\%$ on \emph{sufficient} versus $57.1\%$ on \emph{insufficient} questions, and injecting the gold captions and frames recovers only $+1.4$ points for the former but $+6.5$ for the latter. This suggests that the verdict separates well-supported from under-supported predictions and that \emph{insufficient} cases tend to lack decisive evidence; caption-only oracles ($72.6\%/58.4\%$) further indicate our memory already carries most of the answer signal. A qualitative trace is provided in the supplementary material.

\paragraph{Efficacy of Rubric-Guided Verification.}
To locate where the verifier's $7.8$-point gain arises, we compare per-type accuracy on Video-MME(long) before and after rubric-guided verification. As shown in Figure~\ref{fig:analysis}, the gains concentrate on evidence-sensitive categories: Temporal Reasoning ($+17.6$), Counting ($+14.6$), and Action Recognition ($+12.7$), which hinge on ordering, quantity, or action, while high-baseline gist categories change little. No category regresses; only Spatial Perception, with three questions, is unchanged. This confirms that verification helps the categories where decisive evidence, rather than relevance, determines the answer. These are precisely the categories in which topically relevant segments most often omit the deciding detail, which is why sufficiency verification contributes the single largest jump in the aggregate ablation.

\section{Conclusion}

We presented REVEAL, a framework for long-video question answering that separates evidence sufficiency from retrieval relevance. Rather than using rubrics only to evaluate outputs or shape training, REVEAL brings them into the inference loop: a rubric-guided verifier judges whether the retrieved evidence is sufficient and turns unmet criteria into a missing-evidence analysis that drives targeted retrieval over an offline-online video memory, with the rubrics induced automatically from contrastive pairs of decisive and misleading evidence. Across several long-video benchmarks, REVEAL attains state-of-the-art accuracy, surpassing strong open-source and closed-source MLLMs, and our analyses attribute the gains to explicit sufficiency verification rather than to feeding more relevant evidence. A current limitation is that the online memory does not yet feed verification signals back to continually revise the offline memory, which we leave to future work. More broadly, our results suggest that explicit evidence-sufficiency verification is an interpretable control signal that could extend to other retrieval-based and agent-based multimodal tasks.

\bibliography{references}

@article{fu2024videomme,
  title={Video-{MME}: The First-Ever Comprehensive Evaluation Benchmark of Multi-modal {LLMs} in Video Analysis},
  author={Fu, Chaoyou and Dai, Yuhan and Luo, Yongdong and Li, Lei and Ren, Shuhuai and Zhang, Renrui and Wang, Zihan and Zhou, Chenyu and Shen, Yunhang and Zhang, Mengdan and others},
  journal={arXiv preprint arXiv:2405.21075},
  year={2024}
}

@article{zhou2024mlvu,
  title={{MLVU}: Benchmarking Multi-task Long Video Understanding},
  author={Zhou, Junjie and Shu, Yan and Zhao, Bo and Wu, Boya and Liang, Zhengyang and Xiao, Shitao and Qin, Minghao and Yang, Xi and Xiong, Yongping and Zhang, Bo and others},
  journal={arXiv preprint arXiv:2406.04264},
  year={2024}
}

@article{wu2024longvideobench,
  title={{LongVideoBench}: A Benchmark for Long-context Interleaved Video-Language Understanding},
  author={Wu, Haoning and Li, Dongxu and Chen, Bei and Li, Junnan},
  journal={arXiv preprint arXiv:2407.15754},
  year={2024}
}

@article{wang2024lvbench,
  title={{LVBench}: An Extreme Long Video Understanding Benchmark},
  author={Wang, Weihan and He, Zehai and Hong, Wenyi and Cheng, Yean and Zhang, Xiaohan and Qi, Ji and Gu, Xiaotao and Huang, Shiyu and Xu, Bin and Dong, Yuxiao and others},
  journal={arXiv preprint arXiv:2406.08035},
  year={2024}
}

@article{wang2024videoagentlongform,
  title={{VideoAgent}: Long-form Video Understanding with Large Language Model as Agent},
  author={Wang, Xiaohan and Zhang, Yuhui and Zohar, Orr and Yeung-Levy, Serena},
  journal={arXiv preprint arXiv:2403.10517},
  year={2024}
}

@article{fan2024videoagentmemory,
  title={{VideoAgent}: A Memory-augmented Multimodal Agent for Video Understanding},
  author={Fan, Yue and Ma, Xiaojian and Wu, Rujie and Du, Yuntao and Li, Jiaqi and Gao, Zhi and Li, Qing},
  journal={arXiv preprint arXiv:2403.11481},
  year={2024}
}

@article{yang2024vca,
  title={{VCA}: Video Curious Agent for Long Video Understanding},
  author={Yang, Zeyuan and Chen, Delin and Yu, Xueyang and Shen, Maohao and Gan, Chuang},
  journal={arXiv preprint arXiv:2412.10471},
  year={2024}
}

@article{chen2025lvagent,
  title={{LVAgent}: Long Video Understanding by Multi-Round Dynamical Collaboration of {MLLM} Agents},
  author={Chen, Boyu and Yue, Zhengrong and Chen, Siran and Wang, Zikang and Liu, Yang and Li, Peng and Wang, Yali},
  journal={arXiv preprint arXiv:2503.10200},
  year={2025}
}

@inproceedings{lewis2020rag,
  title={Retrieval-Augmented Generation for Knowledge-Intensive {NLP} Tasks},
  author={Lewis, Patrick and Perez, Ethan and Piktus, Aleksandra and Petroni, Fabio and Karpukhin, Vladimir and Goyal, Naman and K{\"u}ttler, Heinrich and Lewis, Mike and Yih, Wen-tau and Rockt{\"a}schel, Tim and others},
  booktitle={Advances in Neural Information Processing Systems},
  volume={33},
  pages={9459--9474},
  year={2020}
}

@inproceedings{yao2023react,
  title={{ReAct}: Synergizing Reasoning and Acting in Language Models},
  author={Yao, Shunyu and Zhao, Jeffrey and Yu, Dian and Du, Nan and Shafran, Izhak and Narasimhan, Karthik and Cao, Yuan},
  booktitle={International Conference on Learning Representations},
  year={2023}
}

@inproceedings{shinn2023reflexion,
  title={Reflexion: Language Agents with Verbal Reinforcement Learning},
  author={Shinn, Noah and Cassano, Federico and Gopinath, Ashwin and Narasimhan, Karthik and Yao, Shunyu},
  booktitle={Advances in Neural Information Processing Systems},
  year={2023}
}

@inproceedings{liu2023geval,
  title={{G-Eval}: {NLG} Evaluation using {GPT-4} with Better Human Alignment},
  author={Liu, Yang and Iter, Dan and Xu, Yichong and Wang, Shuohang and Xu, Ruochen and Zhu, Chenguang},
  booktitle={Proceedings of the 2023 Conference on Empirical Methods in Natural Language Processing},
  pages={2511--2522},
  year={2023}
}

@inproceedings{zheng2023judging,
  title={Judging {LLM}-as-a-Judge with {MT-Bench} and Chatbot Arena},
  author={Zheng, Lianmin and Chiang, Wei-Lin and Sheng, Ying and Zhuang, Siyuan and Wu, Zhanghao and Zhuang, Yonghao and Lin, Zi and Li, Zhuohan and Li, Dacheng and Xing, Eric P. and others},
  booktitle={Advances in Neural Information Processing Systems},
  year={2023}
}

@article{kim2023prometheus,
  title={Prometheus: Inducing Fine-grained Evaluation Capability in Language Models},
  author={Kim, Seungone and Shin, Jamin and Cho, Yejin and Jang, Joel and Longpre, Shayne and Lee, Hwaran and Yun, Sangdoo and Shin, Seongjin and Kim, Sungdong and Thorne, James and Seo, Minjoon},
  journal={arXiv preprint arXiv:2310.08491},
  year={2023}
}

@inproceedings{zhai2023siglip,
  title={Sigmoid Loss for Language Image Pre-Training},
  author={Zhai, Xiaohua and Mustafa, Basil and Kolesnikov, Alexander and Beyer, Lucas},
  booktitle={Proceedings of the IEEE/CVF International Conference on Computer Vision},
  pages={11975--11986},
  year={2023}
}

@article{chen2024bgem3,
  title={{BGE M3}-Embedding: Multi-Lingual, Multi-Functionality, Multi-Granularity Text Embeddings Through Self-Knowledge Distillation},
  author={Chen, Jianlv and Xiao, Shitao and Zhang, Peitian and Luo, Kun and Lian, Defu and Liu, Zheng},
  journal={arXiv preprint arXiv:2402.03216},
  year={2024}
}

@inproceedings{radford2021clip,
  title={Learning Transferable Visual Models from Natural Language Supervision},
  author={Radford, Alec and Kim, Jong Wook and Hallacy, Chris and Ramesh, Aditya and Goh, Gabriel and Agarwal, Sandhini and Sastry, Girish and Askell, Amanda and Mishkin, Pamela and Clark, Jack and others},
  booktitle={International Conference on Machine Learning},
  pages={8748--8763},
  year={2021}
}

@article{wang2025avp,
  title={Active Video Perception: Iterative Evidence Seeking for Agentic Long Video Understanding},
  author={Wang, Ziyang and Zhou, Honglu and Wang, Shijie and Li, Junnan and Xiong, Caiming and Savarese, Silvio and Bansal, Mohit and Ryoo, Michael S. and Niebles, Juan Carlos},
  journal={arXiv preprint arXiv:2512.05774},
  year={2025}
}

@article{kurpath2025longshotbench,
  title={{LongShOTBench}: A Benchmark and Agentic Framework for Omni-Modal Reasoning and Tool Use in Long Videos},
  author={Kurpath, Mohammed Irfan and Kaithakkodan, Jaseel Muhammad and Zhou, Jinxing and Mullappilly, Sahal Shaji and Almansoori, Mohammad and Ahsan, Noor and Kalmakhanbet, Beknur and Shikhar, Sambal and Lalla, Rishabh and Lahoud, Jean and others},
  journal={arXiv preprint arXiv:2512.16978},
  year={2025}
}

@article{waheed2025videojudge,
  title={{VideoJudge}: Bootstrapping Enables Scalable Supervision of {MLLM}-as-a-Judge for Video Understanding},
  author={Waheed, Abdul and Wu, Zhen and Alharthi, Dareen and Kim, Seungone and Raj, Bhiksha},
  journal={arXiv preprint arXiv:2509.21451},
  year={2025}
}

@article{arora2025healthbench,
  title={{HealthBench}: Evaluating Large Language Models Towards Improved Human Health},
  author={Arora, Rahul K. and Wei, Jason and Hicks, Rebecca Soskin and Bowman, Preston and Qui{\~n}onero-Candela, Joaquin and Tsimpourlas, Foivos and Sharman, Michael and Shah, Meghan and Vallone, Andrea and Beutel, Alex and others},
  journal={arXiv preprint arXiv:2505.08775},
  year={2025}
}

@article{chen2024cgbench,
  title={{CG-Bench}: Clue-grounded Question Answering Benchmark for Long Video Understanding},
  author={Chen, Guo and Liu, Yicheng and Huang, Yifei and He, Yuping and Pei, Baoqi and Xu, Jilan and Wang, Yali and Lu, Tong and Wang, Limin},
  journal={arXiv preprint arXiv:2412.12075},
  year={2024}
}

@article{liu2025videomind,
  title={{VideoMind}: A Chain-of-{LoRA} Agent for Long Video Reasoning},
  author={Liu, Ye and Lin, Kevin Qinghong and Chen, Chang Wen and Shou, Mike Zheng},
  journal={arXiv preprint arXiv:2503.13444},
  year={2025}
}

@article{liu2026rubricssurvey,
  title={The Rules of the Game: A Survey of Rubrics for Large Language Models},
  author={Liu, Wenhan and Jin, Jiajie and Huang, Zhaoheng and Wen, Tongyu and Dong, Guanting and Zhao, Ziliang and Zhu, Yutao and Dou, Zhicheng and Wen, Ji-Rong},
  journal={arXiv preprint},
  year={2026}
}

@article{liu2025openrubrics,
  title={{OpenRubrics}: Towards Scalable Synthetic Rubric Generation for Reward Modeling and {LLM} Alignment},
  author={Liu, Tianci and Xu, Ran and Yu, Tony and Hong, Ilgee and Yang, Carl and Zhao, Tuo and Wang, Haoyu},
  journal={arXiv preprint arXiv:2510.07743},
  year={2025}
}

@article{li2026rubrichub,
  title={{RubricHub}: A Comprehensive and Highly Discriminative Rubric Dataset via Automated Coarse-to-Fine Generation},
  author={Li, Sunzhu and Zhao, Jiale and Wei, Miteto and Ren, Huimin and Zhou, Yang and Yang, Jingwen and Liu, Shunyu and Zhang, Kaike and Chen, Wei},
  journal={arXiv preprint arXiv:2601.08430},
  year={2026}
}

@article{fan2025sedareval,
  title={{SedarEval}: Automated Evaluation using Self-Adaptive Rubrics},
  author={Fan, Zhiyuan and Wang, Weinong and Wu, Xing and Zhang, Debing},
  journal={arXiv preprint arXiv:2501.15595},
  year={2025}
}

@article{gao2026qworld,
  title={{Qworld}: Question-Specific Evaluation Criteria for {LLMs}},
  author={Gao, Shanghua and Su, Yuchang and Sui, Pengwei and Ginder, Curtis and Zitnik, Marinka},
  journal={arXiv preprint arXiv:2603.23522},
  year={2026}
}

@article{qi2026rift,
  title={{RIFT}: A {RubrIc} Failure Mode Taxonomy and Automated Diagnostics},
  author={Qi, Zhengyang and Dickens, Charles and Pham, Derek and Dsouza, Amanda and Parchami, Armin and Sala, Frederic and Varma, Paroma},
  journal={arXiv preprint arXiv:2604.01375},
  year={2026}
}

@article{yang2026healthscore,
  title={{Health-SCORE}: Towards Scalable Rubrics for Improving Health-{LLMs}},
  author={Yang, Zhichao and Janghorbani, Sepehr and Zhang, Dongxu and Han, Jun and Qian, Qian and Ressler, Andrew and Lyng, Gregory D. and Batra, Sanjit Singh and Tillman, Robert E.},
  journal={arXiv preprint arXiv:2601.18706},
  year={2026}
}

@article{schulman2017ppo,
  title={Proximal Policy Optimization Algorithms},
  author={Schulman, John and Wolski, Filip and Dhariwal, Prafulla and Radford, Alec and Klimov, Oleg},
  journal={arXiv preprint arXiv:1707.06347},
  year={2017}
}

@article{yeo2025gcagent,
  title={{GCAgent}: Long-Video Understanding via Schematic and Narrative Episodic Memory},
  author={Yeo, Jeong Hun and Chung, Sangyun and Park, Sungjune and Kim, Dae Hoe and Moon, Jinyoung and Ro, Yong Man},
  journal={arXiv preprint arXiv:2511.12027},
  year={2025}
}

@article{tian2025egor1,
  title={{Ego-R1}: Chain-of-Tool-Thought for Ultra-Long Egocentric Video Reasoning},
  author={Tian, Shulin and Wang, Ruiqi and Guo, Hongming and Wu, Penghao and Dong, Yuhao and Wang, Xiuying and Yang, Jingkang and Zhang, Hao and Zhu, Hongyuan and Liu, Ziwei},
  journal={arXiv preprint arXiv:2506.13654},
  year={2025}
}

@article{video_chatgpt,
  title={Video-{ChatGPT}: Towards Detailed Video Understanding via Large Vision and Language Models},
  author={Maaz, Muhammad and Rasheed, Hanoona and Khan, Salman and Khan, Fahad Shahbaz},
  journal={arXiv preprint arXiv:2306.05424},
  year={2023}
}

@article{video_llava,
  title={Video-{LLaVA}: Learning United Visual Representation by Alignment Before Projection},
  author={Lin, Bin and Ye, Yang and Zhu, Bin and Cui, Jiaxi and Ning, Munan and Jin, Peng and Yuan, Li},
  journal={arXiv preprint arXiv:2311.10122},
  year={2023}
}

@article{qwen2_vl,
  title={{Qwen2-VL}: Enhancing Vision-Language Model's Perception of the World at Any Resolution},
  author={Wang, Peng and Bai, Shuai and Tan, Sinan and Wang, Shijie and Fan, Zhihao and Bai, Jinze and Chen, Keqin and Liu, Xuejing and Wang, Jialin and Ge, Wenbin and others},
  journal={arXiv preprint arXiv:2409.12191},
  year={2024}
}

@article{longvu,
  title={{LongVU}: Spatiotemporal Adaptive Compression for Long Video-Language Understanding},
  author={Shen, Xiaoqian and Xiong, Yunyang and Zhao, Changsheng and Wu, Lemeng and Chen, Jun and Zhu, Chenchen and Liu, Zechun and Xiao, Fanyi and Varadarajan, Balakrishnan and Bordes, Florian and others},
  journal={arXiv preprint arXiv:2410.17434},
  year={2024}
}

@inproceedings{moviechat,
  title={{MovieChat}: From Dense Token to Sparse Memory for Long Video Understanding},
  author={Song, Enxin and Chai, Wenhao and Wang, Guanhong and Zhang, Yucheng and Zhou, Haoyang and Wu, Feiyang and Chi, Xun and Guo, Xun and Ye, Tian and Zhang, Yanting and others},
  booktitle={Proceedings of the IEEE/CVF Conference on Computer Vision and Pattern Recognition},
  year={2024}
}

@inproceedings{ma_lmm,
  title={{MA-LMM}: Memory-Augmented Large Multimodal Model for Long-Term Video Understanding},
  author={He, Bo and Li, Hengduo and Jang, Young Kyun and Jia, Menglin and Cao, Xuefei and Shah, Ashish and Shrivastava, Abhinav and Lim, Ser-Nam},
  booktitle={Proceedings of the IEEE/CVF Conference on Computer Vision and Pattern Recognition},
  year={2024}
}

@article{video_rag,
  title={Video-{RAG}: Visually-aligned Retrieval-Augmented Long Video Comprehension},
  author={Luo, Yongdong and Zheng, Xiawu and Li, Guilin and Yin, Shukang and Lin, Haojia and Fu, Chaoyou and Huang, Jinfa and Ji, Jiayi and Chao, Fei and Luo, Jiebo and Ji, Rongrong},
  journal={arXiv preprint},
  year={2025}
}

@article{worldmm,
  title={{WorldMM}: Dynamic Multimodal Memory Agent for Long Video Reasoning},
  author={Yeo, Woongyeong and Kim, Kangsan and Yoon, Jaehong and Hwang, Sung Ju},
  journal={arXiv preprint arXiv:2512.02425},
  year={2025}
}

@article{zhou2025ruscarl,
  title={Breaking the Exploration Bottleneck: Rubric-Scaffolded Reinforcement Learning for General {LLM} Reasoning},
  author={Zhou, Yang and Li, Sunzhu and Liu, Shunyu and Fang, Wenkai and Zhang, Kongcheng and Zhao, Jiale and Yang, Jingwen and Zhou, Yihe and Lv, Jianwei and Zheng, Tongya and Lu, Hengtong and Chen, Wei and Xie, Yan and Song, Mingli},
  journal={arXiv preprint arXiv:2508.16949},
  year={2025}
}

@inproceedings{zhang2026chasingtail,
  title={Chasing the Tail: Effective Rubric-based Reward Modeling for Large Language Model Post-Training},
  author={Zhang, Junkai and Wang, Zihao and Gui, Lin and Mysore Sathyendra, Swarnashree and Jeong, Jaehwan and Veitch, Victor and Wang, Wei and He, Yunzhong and Liu, Bing and Jin, Lifeng},
  booktitle={International Conference on Learning Representations (ICLR)},
  year={2026}
}

@inproceedings{wang2025videorts,
  title={{Video-RTS}: Rethinking Reinforcement Learning and Test-Time Scaling for Efficient and Enhanced Video Reasoning},
  author={Wang, Ziyang and Yoon, Jaehong and Yu, Shoubin and Islam, Md Mohaiminul and Bertasius, Gedas and Bansal, Mohit},
  booktitle={Proceedings of the 2025 Conference on Empirical Methods in Natural Language Processing},
  year={2025},
  doi={10.18653/v1/2025.emnlp-main.1428}
}

@article{bai2025qwen3vl,
  title={{Qwen3-VL} Technical Report},
  author={Bai, Shuai and Cai, Yuxuan and Chen, Ruizhe and others},
  journal={arXiv preprint arXiv:2511.21631},
  year={2025}
}

@article{qwen2026qwen35,
  title={{Qwen3.5-Omni} Technical Report},
  author={{Qwen Team}},
  journal={arXiv preprint arXiv:2604.15804},
  year={2026}
}

@misc{google2025gemini25,
  title={{Gemini 2.5}: Pushing the Frontier with Advanced Reasoning, Multimodality, Long Context, and Next Generation Agentic Capabilities},
  author={{Google DeepMind}},
  year={2025},
  howpublished={Technical report}
}

@misc{openai2025gpt5,
  title={{GPT-5} System Card},
  author={{OpenAI}},
  year={2025},
  howpublished={System card}
}

@article{gutierrez2025hipporag2,
  title={From {RAG} to Memory: Non-Parametric Continual Learning for Large Language Models},
  author={Guti{\'e}rrez, Bernal Jim{\'e}nez and Shu, Yiheng and Qi, Weijian and Zhou, Sizhe and Su, Yu},
  journal={arXiv preprint arXiv:2502.14802},
  year={2025}
}

@misc{shen2025vgent,
  title={{Vgent}},
  author={Shen and others},
  year={2025},
  howpublished={Technical report}
}

@misc{long2025m3agent,
  title={{M3-Agent}},
  author={Long and others},
  year={2025},
  howpublished={Technical report}
}

@article{zuo2025videolucy,
  title={{VideoLucy}: Deep Memory Backtracking for Long Video Understanding},
  author={Zuo, Jialong and Deng, Yongtai and Kong, Lingdong and Yang, Jingkang and Jin, Rui and Zhang, Yiwei and Sang, Nong and Pan, Liang and Liu, Ziwei and Gao, Changxin},
  journal={arXiv preprint arXiv:2510.12422},
  year={2025}
}

@inproceedings{asai2023selfrag,
  title={Self-{RAG}: Learning to Retrieve, Generate, and Critique through Self-Reflection},
  author={Asai, Akari and Wu, Zeqiu and Wang, Yizhong and Sil, Avirup and Hajishirzi, Hannaneh},
  booktitle={International Conference on Learning Representations},
  year={2024}
}

@inproceedings{jiang2023flare,
  title={Active Retrieval Augmented Generation},
  author={Jiang, Zhengbao and Xu, Frank F. and Gao, Luyu and Sun, Zhiqing and Liu, Qian and Dwivedi-Yu, Jane and Yang, Yiming and Callan, Jamie and Neubig, Graham},
  booktitle={Proceedings of the 2023 Conference on Empirical Methods in Natural Language Processing},
  pages={7969--7992},
  year={2023}
}

@article{yan2024crag,
  title={Corrective Retrieval Augmented Generation},
  author={Yan, Shi-Qi and Gu, Jia-Chen and Zhu, Yun and Ling, Zhen-Hua},
  journal={arXiv preprint arXiv:2401.15884},
  year={2024}
}

% ============================================================
% Appendix / Supplementary Material (merged for the arXiv version)
% ============================================================
\clearpage
\appendix

\section{Additional Details on Datasets}

We evaluate REVEAL on five public long-video QA benchmarks that together span third-person and egocentric footage and a broad range of durations and reasoning demands. All benchmarks are scored with multiple-choice accuracy, and every method is evaluated on exactly the same question set for each benchmark.

\paragraph{Why these benchmarks.}
We select these five benchmarks so that our evaluation jointly stresses the conditions REVEAL is designed for, namely locating and combining sparse decisive evidence across long temporal horizons. They are chosen to vary along three axes. First, in \emph{viewpoint}, Video-MME-L and LVBench are third-person, whereas EgoLifeQA and Ego-R1 Bench are egocentric, where evidence is sparser and tied to fine-grained daily activities; this tests whether our memory and sufficiency verification transfer across perspectives rather than overfitting to one. Second, in \emph{duration}, the videos range from roughly $30$--$60$ minutes (Video-MME-L) to over an hour (LVBench) to week-scale recordings (Ego-R1 Bench), probing scalability as the search space grows. Third, in \emph{reasoning demand}, the benchmarks emphasize different skills, from global gist and event ordering to referring and temporal localization (LongVideoBench) and cross-event evidence chaining (Ego-R1 Bench). All five are public, widely used, and evaluated under the same multiple-choice protocol, which keeps our results comparable to prior work; in addition, Ego-R1 Bench provides target-time annotations that enable the retrieval-free oracle analysis in the main paper. We deliberately avoid short-video or single-clip benchmarks, since they do not exhibit the long-horizon, sparse-evidence retrieval problem that motivates REVEAL.

\paragraph{Evaluation metric.}
Every benchmark we use is posed as multiple-choice question answering, so we report top-1 answer accuracy, i.e., the fraction of questions whose selected option matches the gold option. We adopt accuracy for three reasons. First, it is the metric defined by the original benchmarks and used by prior work, so our results are directly comparable to published baselines without any re-scoring. Second, it is objective and fully reproducible: correctness is an exact match against a single gold option, which avoids the noise and prompt sensitivity of free-form generation metrics or LLM-as-a-judge scoring. This matters for our setting, where questions probe fine-grained temporal, causal, and counting distinctions and a partial-credit metric could obscure whether the decisive evidence was actually retrieved. Third, a single scalar per benchmark lets us compare performance consistently across viewpoint regimes (third-person vs.\ egocentric) and across widely varying video durations. Because the answer options are fixed, accuracy also cleanly isolates the effect of evidence sufficiency, which is our central question, from confounds introduced by answer-generation style.

\paragraph{Video-MME-L.}
We use the Long split of Video-MME, which contains $900$ questions over $300$ videos of roughly $30$--$60$ minutes each drawn from diverse domains. Human-written subtitles are provided; when available, they are temporally aligned to segments during memory construction. This split stresses long-horizon third-person understanding.

\paragraph{LVBench.}
LVBench targets extreme-length third-person videos averaging over an hour. Because several source videos have been removed from their original hosting site, we evaluate on the publicly retrievable subset of $86$ videos with $1{,}267$ questions and run all baselines on this identical subset, so no result is compared against externally reported numbers computed on the full set.

\paragraph{LongVideoBench.}
We use the public validation split of LongVideoBench: $1{,}337$ questions over $753$ videos. Its questions emphasize referring and temporal reasoning, where the model must locate a specific referenced moment rather than rely on global gist.

\paragraph{EgoLifeQA.}
EgoLifeQA is an egocentric daily-life benchmark; we evaluate the single-participant (A1) setting with $500$ questions. Answer evidence is sparse and tied to fine-grained everyday activities spread across very long recordings.

\paragraph{Ego-R1 Bench.}
Ego-R1 Bench evaluates ultra-long egocentric reasoning over week-scale recordings with $300$ questions, many of which require chaining evidence across widely separated moments. It provides \emph{target-time} annotations marking the interval that contains each answer's supporting evidence, which we use only for the retrieval-free oracle analysis in the main paper.

\section{Additional Implementation Details}

\subsection{Baseline Setup}

For fair comparison, we run each modular baseline with its released pipeline but replace its interchangeable backbone with the same Qwen3.5-27B model used by REVEAL, so that accuracy differences reflect retrieval and memory design rather than model capacity. M3-Agent is the exception: it relies on a task-trained checkpoint and changing its backbone would require retraining, so we evaluate its released checkpoint as published. The direct open-source and commercial MLLMs use neither retrieval nor an explicit memory module: following common practice, we sample frames uniformly at $1$~fps and cap the visual input at $768$ frames, uniformly subsampling any longer video down to this budget, and feed the resulting frames together with the question and options. Unless otherwise stated, all other settings follow each method's official specification.

\subsection{Offline and Online Memory}

\paragraph{Offline memory.}
The offline memory $\mathcal{M}$ stores a persistent, multi-scale representation of the entire video that any later question can query. Because an hour-long video cannot be fed to a VLM in full, $\mathcal{M}$ uses similarity-based variable-length segmentation to preserve coherent events and a three-level hierarchy to expose evidence at multiple temporal granularities. It is constructed once, before any question is observed, through the bottom-up process described below.

We first sample video frames at 1 fps and remove low-quality frames before segmentation. Near-black, near-white, and nearly uniform frames are detected using JPEG size and pixel statistics, while blurred frames are filtered using a Laplacian-variance criterion. The remaining frames are encoded using SigLIP. Rather than partitioning the video into fixed-duration chunks, adjacent frames are greedily grouped into variable-length segments when their visual cosine similarity exceeds a threshold $\theta$, which is set to $0.8$ by default. This similarity-based grouping reduces the likelihood that a coherent action or scene transition is divided across unrelated memory units.

For each segment, the same Qwen3.5-27B video-language model that later serves as planner, verifier, and answerer generates a fine-grained textual description using the segment frames together with temporally aligned subtitles when the dataset provides them; we do not run automatic speech recognition. The captioning prompt preserves answer-relevant information, including identities, objects, actions, state changes, temporal transitions, spatial relations, and counts. Each L1 memory item is represented as
\begin{equation}
m_i^{(1)} = \left(t_i^s, t_i^e, x_i^{(1)}, \mathbf{h}_i^{(1)}, \mathbf{v}_i\right),
\end{equation}
where $(t_i^s, t_i^e)$ denote the temporal boundaries, $x_i^{(1)}$ is the segment-level narrative, $\mathbf{h}_i^{(1)}$ is its BGE-M3 textual embedding, and $\mathbf{v}_i$ is the SigLIP visual embedding of a representative keyframe. Thus, $\mathbf{h}_i^{(1)}$ serves the text-retrieval channel, while $\mathbf{v}_i$ serves the visual channel. The resulting L1 segments are approximately $10$ seconds long on average.

Higher-level memories are constructed bottom-up by summarizing consecutive lower-level entries. Every 12 L1 descriptions are summarized into an L2 event-level representation:
\begin{equation}
x_j^{(2)} = S_2\left(x_{12j}^{(1)}, \ldots, x_{12j+11}^{(1)}\right),
\end{equation}
where $S_2$ is an LLM summarizer instructed to preserve event progression, temporal order, entity interactions, and state transitions. Each L2 entry contains an approximately 120-word event-timeline summary.

Similarly, every 12 consecutive L2 entries are summarized into an L3 video-level representation:
\begin{equation}
x_k^{(3)} = S_3\left(x_{12k}^{(2)}, \ldots, x_{12k+11}^{(2)}\right),
\end{equation}
where $S_3$ produces an approximately 300-word description of the global narrative, major events, and coarse temporal structure. L1 retains local visual details, L2 captures event processes, and L3 provides whole-video context. Because upper levels are constructed from lower-level descriptions rather than by repeatedly processing raw frames, the hierarchy remains consistent with the original segment observations while avoiding additional visual encoding cost.

Each narrative representation $x_i^{(\ell)}$ at level $\ell \in \{1,2,3\}$ is encoded using BGE-M3:
\begin{equation}
\mathbf{h}_i^{(\ell)} = f_{\mathrm{text}}\left(x_i^{(\ell)}\right).
\end{equation}
Accordingly, every memory level stores its narrative together with $\mathbf{h}_i^{(\ell)}$, whereas the visual embedding $\mathbf{v}_i$ is retained only for L1 segments, where it remains temporally aligned with the source frames. The offline memory is therefore
\begin{equation}
\mathcal{M} = \mathcal{M}_{\mathrm{L1}} \cup \mathcal{M}_{\mathrm{L2}} \cup \mathcal{M}_{\mathrm{L3}}.
\end{equation}
The offline memory exposes two independent retrieval channels rather than a single fused index. The narrative embeddings form the primary channel: a text query is matched against them and returns the corresponding narratives across all three levels. The L1 visual embeddings form a separate channel: the memory stores only the visual vectors, and when a query is matched against them, the frames of each retrieved segment are extracted on demand from the original video by its temporal boundaries $(t_i^s, t_i^e)$ and supplied to the answerer when fine-grained visual appearance is required. The two channels are queried separately, so text retrieval returns narratives while visual retrieval returns raw frames. LongVideoBench videos vary widely in length, so we route the frames supplied to the answerer by video duration: for videos longer than $10$~minutes we sample frames densely within the retrieved span, whereas videos of at most $10$~minutes are short enough to be covered in full and are instead sampled uniformly at $1$~fps over their entire duration. Memory construction and rubric-guided retrieval are identical in both cases; only the frames fed to the answerer differ.

\paragraph{Online memory.}
The implementation maintains one \texttt{OnlineMemory} object per question. Its persistent state after round $t$ is
\begin{equation}
\mathcal{O}_t = \left(\mathcal{E}_t, \mathcal{V}_t, \mathcal{H}_t, \mathcal{N}_t, \mathcal{A}_t\right),
\end{equation}
where $\mathcal{E}_t$ is the admitted evidence cache (text, chunk ID, source, and round), $\mathcal{V}_t$ is the set of visited memory IDs, $\mathcal{H}_t$ is the ordered query history, $\mathcal{N}_t$ records whether each attempt contributed new evidence, and $\mathcal{A}_t$ is the verifier's latest missing-evidence analysis. New evidence is rejected when its embedding cosine similarity to the cache exceeds $0.9$. For planner feedback, the implementation exposes only $(\mathcal{H}_t, \mathcal{N}_t, \mathcal{A}_t)$; criterion scores and verdict metadata remain in the execution trace. This distinction prevents verbose verifier state from diluting the next retrieval target.

\subsection{Prompt Details}

\paragraph{Planner prompt.}
Round 0 uses a separate decomposer prompt that requests 2--4 self-contained subquestions of at most 15 words, with one query per option for negation questions. From round 1 onward, the system prompt presents exactly two actions: \texttt{Text Retrieval}, with 1--4 focused queries, and \texttt{Uniform Sampling}, with an empty query list. The user message contains the question and options, prior queries under ``Already tried'', current retrieval-quality records, timestamps covered by current evidence, and $\mathcal{A}_t$ under \texttt{\textless missing\_evidence\textgreater}. The required output is
\begin{quote}
\footnotesize\ttfamily\raggedright
\{"strategy": "Text Retrieval" | "Uniform Sampling",\\
\phantom{x}"queries": ["..."],\\
\phantom{x}"reasoning": "one sentence"\}
\end{quote}
Uniform sampling is instructed for global summaries, events distributed across the video, chronological questions, or repeatedly failed targeted retrieval. Generated queries are additionally rejected when Jaccard similarity to history is at least $0.9$ or BGE cosine similarity to accumulated evidence is at least $0.70$; if no query survives, the executor falls back to uniform sampling. For each surviving query, retrieval takes the top $8$ L1 segment narratives, $4$ L2 event summaries, and $2$ L3 global summaries; the candidates are merged across queries and rubric-reranked to $8$ evidence pieces before being admitted to the cache. The loop runs for at most $K=3$ rounds. These rules match \texttt{src/agents/planner.py} in the evaluated mainline; time and image actions are not enabled.

\paragraph{Verifier prompts.}
Verification uses two calls. The first receives only the question, options, and criterion catalog and returns the names of relevant criteria; its result is cached per question. The second receives the selected rubric and accumulated evidence. For every option and criterion it outputs a sufficiency score of $1$, $0.5$, or $0$, where evidence that clearly rules out an option is as sufficient as evidence that confirms it. Its strict output schema is
\begin{quote}
\footnotesize\ttfamily\raggedright
\{"rubric\_criteria\_scores": \{\\
\phantom{xx}"A": \{"criterion\_name": 0.0\},\\
\phantom{xx}"B": \{"criterion\_name": 0.0\}\},\\
\phantom{x}"unknown\_options": ["A"],\\
\phantom{x}"missing\_evidence\_analysis":\\
\phantom{xx}"specific fact to retrieve"\}
\end{quote}
The model does not emit the final verdict. The implementation computes $\rho_{t,o} = \sum_k w_k z_{t,o,k} / \sum_k w_k$, aggregates $\rho_t = \max_o \rho_{t,o}$, and compares it with the rubric-pool threshold $\tau = 0.7$. Only when the result is insufficient is the nonempty missing-evidence analysis fed back to the planner. This procedure matches \texttt{src/agents/verifier.py} and the \texttt{rubric\_pool.yaml} configuration used by the mainline runs.

\subsection{Contrastive Rubric Construction Details}

\paragraph{Evidence sampling and labeling.}
For each development question, candidate evidence is retrieved using L1@16, L2@16, and all L3 summaries. Given the question, options, and gold answer, a teacher labels a sufficient set $E^+$ containing the decisive information and a misleading set $E^-$ that is topically relevant but omits the critical action, transition, relation, causal clue, or exclusion evidence.

\paragraph{Pair filtering.}
We retain a question only when the answer model is correct with $E^+$, incorrect without video evidence, and incorrect or unsupported with $E^-$. Each retained question yields one positive and one negative evidence chain. This removes prior-answerable and weakly contrastive cases; on Video-MME, approximately 44\% of candidates are filtered out.

\paragraph{Rubric refinement and verification.}
For every retained pair, a proposer may add, rewrite, decompose, delete, or reweight criteria under the constraint that each criterion separates the sufficient and misleading chains. Proposed criteria are clustered, deduplicated, and weight-normalized. The updated library is then evaluated on held-out pairs; misjudged examples form the next working set, and refinement repeats until validation stabilizes. At inference time, the verifier selects relevant criteria from the merged library rather than constructing a new rubric.

As an example, a counting rubric checks whether the evidence enumerates distinct instances, identifies each instance by a defining attribute, and consistently excludes alternative counts. Each criterion supplies explicit boundaries for scores $1$, $0.5$, and $0$, with the weighted score $\rho = \sum_k w_k z_k / \sum_k w_k$ compared against $\tau$.

\subsection{Hyperparameter Selection}

The main hyperparameters were tuned on Video-MME-L during development and then applied unchanged to all other benchmarks. We grid-searched the number of retrieval rounds $K \in \{1,2,3,4,5\}$ and the sufficiency threshold $\tau \in \{0.6, 0.7, 0.8, 0.85, 0.9\}$, and additionally varied the segmentation similarity $\theta$ around its default of $0.8$. For each configuration we compared development accuracy under a fixed per-round retrieval budget and selected the setting that maximized accuracy without incurring unnecessary retrieval, giving $K=3$, $\tau=0.7$, and $\theta=0.8$.

\subsection{Computing Infrastructure}

All experiments were run on a Linux server (Ubuntu~24.04, kernel~6.17) with a 192-core Intel Xeon Platinum 8558 CPU, $818$~GB of RAM, and NVIDIA RTX PRO 6000 (Blackwell, $96$~GB) GPUs (driver~580, CUDA~13). A single $96$~GB GPU is sufficient to host Qwen3.5-27B, which we serve with vLLM~0.20; the number of GPUs used varies across runs and only parallelizes evaluation without affecting results. The retrieval encoders SigLIP-Large and BGE-M3 run under Python~3.11 with PyTorch~2.11 (CUDA~12.8 build), transformers~5.7, sentence-transformers~5.4, and FlagEmbedding~1.4.

\subsection{Efficiency and Cost}

Beyond accuracy, we compare REVEAL's end-to-end cost against representative retrieval- and memory-based baselines on two benchmarks. Table~\ref{tab:efficiency} reports wall-clock runtime, GPU-hours, and total token consumption, and Table~\ref{tab:reveal_rounds_cost} summarizes REVEAL's average retrieval rounds and LLM call cost across all five evaluation benchmarks. For each benchmark, all methods are run on the same question set and the same hardware; all figures are approximate and denoted by ``$\sim$''. REVEAL achieves the highest accuracy among the compared methods while maintaining moderate end-to-end cost: its runtime is below the heavier memory-based baseline and remains within the same order of magnitude as lightweight retrieval systems.

\begin{center}
\begin{minipage}{\columnwidth}
\centering
\footnotesize
\setlength{\tabcolsep}{3.5pt}
\renewcommand{\arraystretch}{1.08}
\begin{tabular*}{\linewidth}{@{\extracolsep{\fill}}lccc@{}}
\toprule
\multicolumn{4}{c}{\textbf{LVBench}} \\
\cmidrule(lr){1-4}
Method & Time (h) & GPU-h & Tok.\ (M) \\
\midrule
\textbf{REVEAL (ours)} & $\sim$5.5 & $\sim$88 & $\sim$145 \\
VideoLucy & $\sim$14 & $\sim$224 & $\sim$230 \\
VideoRAG & $\sim$6.4 & $\sim$102 & $\sim$80 \\
HippoRAG & $\sim$1.55 & $\sim$25 & $\sim$55.9 \\
Vgent & $\sim$3.1 & $\sim$50 & $\sim$97 \\
\addlinespace[4pt]
\multicolumn{4}{c}{\textbf{Video-MME-L}} \\
\cmidrule(lr){1-4}
Method & Time (h) & GPU-h & Tok.\ (M) \\
\midrule
\textbf{REVEAL (ours)} & $\sim$5.5 & $\sim$88 & $\sim$199 \\
VideoLucy & $\sim$10 & $\sim$160 & $\sim$164 \\
VideoRAG & $\sim$8 & $\sim$128 & $\sim$102 \\
HippoRAG & $\sim$3.33 & $\sim$53 & $\sim$122.5 \\
Vgent & $\sim$6.4 & $\sim$102 & $\sim$136.75 \\
\bottomrule
\end{tabular*}
\captionof{table}{End-to-end efficiency on LVBench and Video-MME-L. All methods use the same question set and hardware within each benchmark; values are approximate ($\sim$).}
\label{tab:efficiency}
\end{minipage}
\end{center}

\begin{center}
\begin{minipage}{\columnwidth}
\centering
\scriptsize
\setlength{\tabcolsep}{2.5pt}
\renewcommand{\arraystretch}{1.08}
\begin{tabular*}{\linewidth}{@{\extracolsep{\fill}}lccc@{}}
\toprule
Dataset & Iter dist. (1/2/3) & Avg. rounds & LLM calls/Q \\
\midrule
Video-MME-L & 589 / 48 / 263 & 1.64 & 5.3 \\
LVBench & 407 / 64 / 796 & 2.31 & 6.6 \\
LongVideoBench & 28 / 16 / 90 & 2.46 & 6.9 \\
EgoLife (A1) & 19 / 1 / 30 & 2.22 & 6.4 \\
EgoR1 (6 users) & 56 / 12 / 82 & 2.17 & 6.3 \\
\bottomrule
\end{tabular*}
\captionof{table}{Retrieval rounds and LLM call cost of REVEAL across five benchmarks. ``Iter dist.'' gives the number of questions that terminate after 1, 2, or 3 retrieval rounds.}
\label{tab:reveal_rounds_cost}
\end{minipage}
\end{center}

\paragraph{Why token count and runtime are not proportional.}
The Video-MME-L result illustrates that total token volume is not the sole determinant of wall-clock time: REVEAL processes approximately $199$M tokens yet completes faster than several methods with lower token counts. Token count measures aggregate model input and output, whereas runtime also depends on the number of sequential calls, batching efficiency, repeated retrieval, and autoregressive decoding. REVEAL reduces the critical path in several ways. Its hierarchical memory is constructed once per video and reused across questions, and its rubric pool is fixed rather than regenerated for every query. During online QA, retrieval is capped at $K=3$ rounds, stops as soon as the accumulated evidence is sufficient, and uses online memory to reject duplicate evidence and avoid repeating unsuccessful searches. The system averages only 1.64--2.46 retrieval rounds and 5.3--6.9 LLM calls per question across the five benchmarks (Table~\ref{tab:reveal_rounds_cost}); on the two efficiency benchmarks specifically, it uses 5.3 calls per question on Video-MME-L and 6.6 on LVBench. Thus, the additional tokens support reusable multi-scale memory and explicit verification, while the number of sequential dependencies remains bounded; this allows substantial token volume to be processed without a proportional increase in end-to-end latency.

\paragraph{Accuracy--efficiency trade-off.}
Taken together with the main results, Table~\ref{tab:efficiency} shows that REVEAL's consistent accuracy advantage does not require a proportional increase in runtime. REVEAL is both more accurate and faster than VideoLucy and VideoRAG on both benchmarks, and it is also faster than Vgent on Video-MME-L. HippoRAG, and Vgent on LVBench, provide lower wall-clock cost, but at the expense of lower accuracy. This favorable trade-off follows from bounded and selective computation: verification adds further retrieval only when the current evidence is insufficient, and each repair round targets a diagnosed evidence gap rather than broadly expanding the context. Online memory additionally suppresses redundant retrieval, while early stopping avoids unnecessary rounds once decisive evidence has been collected. REVEAL therefore spends its additional computation on improving evidence quality rather than simply increasing evidence volume, yielding higher accuracy with controlled runtime overhead. Table~\ref{tab:reveal_rounds_cost} further shows that this control extends beyond the two efficiency benchmarks: across all five evaluation sets, REVEAL typically stops after roughly two retrieval rounds on average and stays within a narrow band of 5.3--6.9 LLM calls per question, indicating that the retrieve--verify--repair loop remains stable rather than expanding into many serial steps.

\section{Additional Description on Experiments}

This section provides additional qualitative support for the Analysis in the main paper, complementing the evidence-volume, verifier-calibration, and per-type breakdowns reported there.

\subsection{Case Study}

Figure~\ref{fig:appendix_case} traces one Video-MME-L question that the no-rubric variant answers incorrectly and REVEAL answers correctly. The question asks which listed event is \emph{not} shown, and the first retrieval round returns segments that are all topically on-video, so a relevance-only reader commits to a plausible-but-wrong option. The rubric verifier instead marks every option \texttt{unknown} and returns \emph{insufficient}: no segment yet grounds the presence or absence of the decisive event. The planner turns this diagnosis into a targeted repair query, retrieves the two segments that settle it, and the verifier then returns \emph{sufficient}, after which the answerer selects the correct option. This is the intended behaviour---separating relevance from sufficiency and repairing the specific gap---made concrete.

\begin{center}
\begin{minipage}{\columnwidth}
\centering
\begin{tabular}{@{}p{0.94\columnwidth}@{}}
\toprule
\textbf{Q.} Which of the following is \emph{not} depicted in the video? \\[2pt]
\quad(A) Leopard climbs trees \quad (B) Hyenas prey on animals \\
\quad(C) A lion is attacked by lions \\
\quad\textbf{(D) Eagles eat animal bodies} $\leftarrow$ gold \\
\midrule
\textbf{Round 0}\quad decompose into four per-option probes; retrieve 26 segments. \newline
\emph{Verifier:} all options \texttt{unknown} $\Rightarrow$ \emph{insufficient}. \\[3pt]
\textbf{Round 1}\quad repair query: \emph{``visual evidence of eagles eating animal carcasses''}; $+2$ decisive segments. \newline
\emph{Verifier:} \emph{sufficient}. \\
\midrule
No-rubric variant $\rightarrow$ \textbf{(C)} ($\times$) \qquad REVEAL $\rightarrow$ \textbf{(D)} ($\checkmark$) \\
\bottomrule
\end{tabular}
\captionof{figure}{A representative retrieve--verify--repair trace (Video-MME-L). Relevance-only retrieval is insufficient for a negation question; the rubric verifier localizes the gap and one targeted repair round supplies the decisive evidence, flipping the answer from wrong to right.}
\label{fig:appendix_case}
\end{minipage}
\end{center}

\end{document}